\documentclass[11pt]{article}

\usepackage[utf8]{inputenc}
\usepackage[T1]{fontenc}
\usepackage{lmodern}
\usepackage{microtype}
\usepackage[margin=1in]{geometry}
\usepackage{amsmath,amssymb,amsfonts}
\usepackage{mathtools}
\usepackage{bm}
\usepackage{graphicx}
\usepackage{booktabs}
\usepackage{multirow}
\usepackage{xcolor}
\usepackage{listings}
\usepackage{caption}
\usepackage{subcaption}
\usepackage[numbers,sort&compress]{natbib}
\usepackage[colorlinks=true,linkcolor=blue!60!black,citecolor=blue!60!black,urlcolor=blue!60!black]{hyperref}
\usepackage{url}
\usepackage{enumitem}
\usepackage{placeins}

\definecolor{codebg}{RGB}{247,247,247}
\definecolor{codekw}{RGB}{0,90,160}
\definecolor{codecm}{RGB}{100,120,100}
\newcommand{\Ltop}{L_0}
\newcommand{\firedfrac}{\texttt{fired\_frac}}

\title{From Geometric Recovery to Causal Validation:\\
A Reproducible Audit of Sparse Autoencoder Features,\\
from Superposition Geometry to Causal Inertness}

\author{
  Mohamed Abdessalem Bal\\
  Independent Researcher, Algiers, Algeria\\
  \texttt{mohamed.bal@devupdz.com}\\[2pt]
  \small Code: \url{https://github.com/mohamed-bal/superposition-to-monosemanticity}\\
  \small \phantom{Code: }\url{https://github.com/mohamed-bal/sae-causal-audit}
}

\date{July 2026}

\begin{document}
\maketitle

\begin{abstract}
Sparse autoencoders (SAEs) are the de facto standard for decomposing superposed neural representations into interpretable features, and the field's evaluation practice relies predominantly on correlational recovery metrics---cosine similarity between ground-truth (or probed) directions and learned decoder atoms. We show, in a fully controlled setting where ground truth is known, that this practice conflates two distinct empirical claims: decoder-geometry alignment and encoder-activation behavior. We first reproduce, from scratch and with multi-seed statistics, the superposition phase diagram of Elhage et al.\ (2022), identifying along the way a convergence artifact at high sparsity and a previously under-described \emph{diffuse sharing} regime at extreme overcompleteness. We then reproduce the TopK-versus-$L_1$ SAE comparison of Gao et al.\ (2024), including direct quantitative evidence of $L_1$ shrinkage (activation-magnitude refinement recovers 91.2\% of the $L_1$ SAE's reconstruction gap while shifting magnitudes $+22.5$\%, versus $-0.05$\% for TopK). Our central result is causal: subjecting every correlationally recovered feature to ablation and steering interventions, we find that up to 77\% of features passing a standard recovery bar (cosine $\ge 0.90$) in a degraded SAE---and 9\% in a well-trained one---are \emph{causally inert}: the matched atom never fires when the feature is present, including matches at cosine $\approx 1.000$. We package the methodology as \texttt{sae-causal-audit}, a model-agnostic, protocol-typed instrument with a hash-verified deterministic reproduction pipeline, and re-auditing the setting under it refines the finding twice: causal inertness decomposes by \emph{cause} into structural inertness (traceable to antipodal-pair superposition geometry and present in good SAEs) and competitive inertness (a TopK-selection pathology of degraded SAEs), and by \emph{direction} into read-inertness and write-inertness, which five antipodal pairs dissociate completely---features that are unmonitorable yet steerable through the same atom, with steering specificities of 143--310 attached to ablation effects of exactly zero. We additionally document why byte-exact cross-platform reproducibility is unavailable by construction for this class of workload, and propose reporting reproducibility as a stack of claims with explicit scopes. Applying the instrument to a published production SAE (GPT-2-small, 83 concepts, one hook layer) reproduces the qualitative pattern at small scale (14\% causally inert among recovered features) and surfaces an unanticipated real-model signal: a handful of decoder atoms recur as the nearest correlational match for dozens of semantically unrelated concepts, replicated across three independently constructed concept batches---real-model evidence for the same dictionary under-splitting the toy antipodal-pair mechanism was built to make legible.
\end{abstract}

% ==================================================================
\section{Introduction}
% ==================================================================

Mechanistic interpretability's current toolkit rests on a specific pipeline: hypothesize that networks represent more features than they have dimensions by packing them into almost-orthogonal directions (\emph{superposition}~\cite{elhage2022toy}), train a sparse autoencoder to un-mix those directions into an overcomplete dictionary~\cite{bricken2023monosemanticity}, and evaluate the dictionary by how well its atoms align, geometrically, with known or probed feature directions. This pipeline scales: Anthropic extracted millions of features from a production model~\cite{templeton2024scaling}, and OpenAI and Google DeepMind have published comparable open-weight SAE suites~\cite{gao2024scaling,lieberum2024gemmascope}.

Recovering \emph{a} sparse, plausible-looking decomposition, however, is not the same as demonstrating that the decomposition is the one the model actually uses. No interpretability method today gives a certified, complete account of a model's internal computation~\cite{bereska2024mechanistic}. The evaluation gap this paper targets is narrower and precisely characterizable: the field's dominant recovery metric---unsigned cosine similarity between a target direction and its best-matching \emph{decoder} atom---measures reconstruction geometry, while whether an atom ever \emph{fires} for a given input is governed by the \emph{encoder} and its selection nonlinearity (TopK, ReLU). These are two different empirical claims, and only an intervention can tell them apart.

We work in the most favorable setting that can be constructed for the recovery pipeline: a small synthetic model in which ground truth is fully known and controlled. This choice is deliberate. The toy setting is what makes rigorous causal validation \emph{possible}; it is also exactly what makes every absolute number reported here a calibration point for the easy case, not a benchmark for a production model.

\paragraph{Contributions.}
\begin{enumerate}[leftmargin=2em,itemsep=2pt]
  \item \textbf{A multi-seed reproduction of the superposition phase diagram}~\cite{elhage2022toy}, with two additions: a documented convergence artifact at high sparsity (under-training masquerading as superposition, resolved by sparsity-aware step budgets), and an under-described third geometric regime---\emph{diffuse sharing}---at extreme overcompleteness, in which every feature is substantially represented ($\lVert W_i\rVert^2 \approx 1.40$) while feature dimensionality reads $D_i \approx 0.08$ (Section~\ref{sec:phase}).
  \item \textbf{A from-scratch reproduction of the TopK-vs-$L_1$ Pareto comparison}~\cite{gao2024scaling}, including a direct activation-refinement test that quantifies $L_1$ shrinkage~\cite{tibshirani1996} rather than asserting it (Section~\ref{sec:sae}).
  \item \textbf{A causal validation battery} (ablation and sign-correct steering, propagated through the model's actual output stage) applied to every correlationally recovered feature, establishing that correlational recovery and causal efficacy diverge by a measured margin: 17 of 22 matched features causally inert in a degraded SAE, 2 of 22 in a well-trained one, with inert matches at cosine up to $0.9998$ (Sections~\ref{sec:causal} and~\ref{sec:reaudit}).
  \item \textbf{\texttt{sae-causal-audit}}, a model-agnostic instrument packaging the battery behind two structural protocols (any object that can \texttt{encode}, \texttt{decode}, and expose its dictionary is auditable, including \texttt{sae\_lens.SAE}), with interface design that makes a documented sign-handling bug class unrepresentable (Section~\ref{sec:instrument}).
  \item \textbf{A two-axis taxonomy of causal inertness.} By \emph{cause}: structural inertness, generated systematically by antipodal-pair superposition geometry under a positive-pass encoder and \emph{not} removed by better SAE training; versus competitive inertness, a TopK-competition pathology of degraded dictionaries that is simultaneously the locus of numerical boundary instability. By \emph{direction}: read-inertness (ablation-based monitoring is blind) versus write-inertness (steering is impotent), which antipodal pairs dissociate completely (Section~\ref{sec:anatomy}).
  \item \textbf{Reproducibility as a stack of claims.} We document, through a three-layer debugging episode on CI infrastructure, why byte-exact reproducibility across platforms is unavailable by construction for floating-point training workloads, and we propose---and implement---a two-tier guarantee: SHA-256 byte-exactness within one pinned environment, semantic equality ($\mathrm{rtol}=10^{-4}$, boundary-sensitive integer counts within $\pm 1$) everywhere (Section~\ref{sec:repro}).
  \item \textbf{A first real-model census and an unanticipated atom-collision finding.} Auditing a published production SAE (\texttt{gpt2-small-res-jb}) against 83 hand-authored concepts spanning unrelated semantic domains recovers 7 correlationally matched features, 1 (14\%) causally inert---small-scale but qualitatively consistent with the toy regime. A small number of decoder atoms recur as the nearest match across dozens of semantically disjoint concepts, replicated across three independently constructed concept batches of growing size and diversity, and independently ruled out as a prompt-template artifact by a controlled negative case (Section~\ref{sec:real}).
\end{enumerate}

% ==================================================================
\section{Background and Related Work}
% ==================================================================

\paragraph{Superposition.}
Individual neurons in trained networks frequently activate for several unrelated concepts---\emph{polysemanticity}---which is why reading meaning off single neurons has never worked reliably as a general strategy. Elhage et al.~\cite{elhage2022toy} propose a two-part explanation: the \emph{linear representation hypothesis} (features correspond to directions in activation space, not basis coordinates) and the \emph{superposition hypothesis} (when features are sparse---rarely simultaneously active---a model benefits from representing more features than dimensions by assigning them almost-orthogonal directions, accepting rare interference in exchange for capacity). The geometric justification is the Johnson--Lindenstrauss phenomenon~\cite{johnson1984}: a $d$-dimensional space admits exponentially more than $d$ near-orthogonal directions once ``exactly orthogonal'' is relaxed to ``inner product bounded near zero.'' Under symmetric two-feature competition for one dimension, the stable configuration found by gradient descent is the \emph{antipodal pair} $W_i = -W_j$; with more features, higher-order polytope-like arrangements emerge, structurally analogous to the Thomson problem of packing mutually repelling charges on a sphere. This paper reproduces the two-feature antipodal case exactly and the large-scale phase structure statistically; full polytope geometry (3--8 feature models in isolation) is out of scope.

\paragraph{Dictionary learning and SAEs.}
Recovering individual features from a superposed mixture is a sparse coding problem with a history predating the interpretability boom: Olshausen and Field~\cite{olshausen1996} explained simple-cell receptive fields as emergent sparse codes over natural images. The interpretability instantiation~\cite{bricken2023monosemanticity,templeton2024scaling} trains the same mathematical object on transformer activations. Exact $\Ltop$ (cardinality) minimization is NP-hard; the classical convex relaxation penalizes the code's $L_1$ norm, at the documented cost of \emph{shrinkage}~\cite{tibshirani1996}: every active coefficient is biased toward zero, correct ones included. Gao et al.~\cite{gao2024scaling} remove the penalty entirely and enforce sparsity structurally via a TopK selection in the encoder, with an auxiliary loss resurrecting ``dead'' atoms; subsequent variants include Gated SAEs~\cite{rajamanoharan2024gated} and BatchTopK~\cite{bussmann2024batchtopk}. Published open-weight suites now exist for GPT-2 and Gemma-2~\cite{gao2024scaling,lieberum2024gemmascope}.

\paragraph{Evaluation practice and its gap.}
Recovery is conventionally scored correlationally: a ground-truth (or linear-probe) direction counts as recovered if its best-matching decoder atom exceeds a cosine-similarity bar, typically taken unsigned because a feature and its exact negation are equally good correlational matches. Causal evaluation---ablation, activation patching, steering---is standard for \emph{circuits} but is not routinely applied per-feature to certify recovery claims. A 2024 survey~\cite{bereska2024mechanistic} additionally flags an unresolved concern: increasingly capable models could develop internal structure that resists or misleads the very techniques probing them. Our results bear on this indirectly (Section~\ref{sec:discussion}): if decoder geometry can diverge this far from encoder behavior \emph{without any adversarial pressure}, purely as an ordinary training-dynamics artifact, the space of ``geometrically plausible, causally misleading'' solutions available under actual optimization pressure is at least as large.

\paragraph{Firing failures and proxy-metric gaps.}
The closest prior work to our central finding is \emph{feature absorption}~\cite{chanin2024absorption}: on production LLM SAEs, Chanin et al.\ show that seemingly monosemantic latents fail to fire on inputs where their concept is plainly present, the activation having been ``absorbed'' into more specific latents---a failure mode caused by the sparsity objective whenever the underlying features form a hierarchy, and not resolved by varying SAE size or sparsity. The observable symptom---an apparently recovered latent that does not activate when its feature occurs---is the same one this paper measures, and the two mechanisms are complementary rather than competing. Absorption requires hierarchical feature structure; the \emph{structural inertness} we isolate (Section~\ref{sec:anatomy}) arises with no hierarchy at all, from antipodal-pair superposition geometry meeting a positive-pass encoder, and our \emph{competitive inertness} is a TopK-selection pathology of degraded dictionaries. Our setting also differs in what it can certify: absorption is detected on real models via probing-based metrics, necessarily correlational at the point of measurement, whereas our audit is interventional end to end (ablation and signed steering propagated through the model's output) against exact ground truth---which is what exposes the read/write dissociation of Section~\ref{sec:taxonomy}, a decomposition invisible to firing statistics alone, since an atom that never fires for a feature can still steer it with high specificity. On the evaluation-practice side, SAEBench~\cite{karvonen2025saebench} documents at benchmark scale that gains on unsupervised proxy metrics do not reliably translate into practical performance across eight evaluation axes and 200+ SAEs; the present work contributes the per-feature, per-direction causal instrument that this broader diagnosis calls for.

% ==================================================================
\section{Experimental Setup}
\label{sec:setup}
% ==================================================================

\subsection{Toy model}
All experiments build on a small, fully from-scratch model (no pretrained weights, no external data; seconds on CPU):
\begin{align}
h &= W x, & \hat{x} &= \mathrm{ReLU}(W^{\top} h + b),
\end{align}
with $W \in \mathbb{R}^{n_{\mathrm{hidden}} \times n_{\mathrm{features}}}$ and the interesting regime $n_{\mathrm{hidden}} < n_{\mathrm{features}}$: a bottleneck that forces a choice about what to represent. Inputs are synthetic and sparse: each of the $n_{\mathrm{features}}$ entries is independently zero with probability $s$ (the \emph{sparsity}) and drawn $\mathrm{Uniform}(0,1)$ otherwise. Training minimizes an importance-weighted reconstruction loss
\begin{equation}
\mathcal{L} = \sum_i I_i \,(x_i - \hat{x}_i)^2 .
\end{equation}

Two importance schedules answer different questions and are never mixed within an experiment:
\begin{itemize}[leftmargin=1.6em,itemsep=1pt]
  \item \textbf{Decayed importance} ($I_i = 0.9^{\,i}$) breaks symmetry so the model has a principled reason to prioritize. It produces a sharp emergent-triage finding of its own: training 32 features into an 8-dimensional bottleneck at $s=0.95$, the optimizer does not spread capacity thinly---it fully drops the 10 lowest-importance features (columns of $W$ collapse to near zero) and represents the remaining 22 well. Nothing hand-codes which features are dropped. This configuration (``32$\to$8, $s=0.95$, decayed'') is the substrate for every SAE experiment below.
  \item \textbf{Uniform importance} ($I_i = 1$) removes prioritization as a confound and is used exclusively for the phase diagram, isolating pure geometric structure.
\end{itemize}

\subsection{Measuring superposition: feature dimensionality and its blind spot}
\label{sec:featdim}
Following Elhage et al.~\cite{elhage2022toy}, each feature's effective ownership of a dedicated dimension is
\begin{equation}
D_i = \frac{\lVert W_i \rVert^2}{\sum_j (W_i \cdot W_j)^2}.
\end{equation}
Three exact cases, each enforced by a unit test rather than asserted: orthonormal features give $D_i = 1$; an antipodal pair gives $D_i = 0.5$ for both members; and---a subtlety with consequences for a real finding below---$D_i$ is a \emph{ratio} of squared norms and therefore not a magnitude detector. For $W = \begin{psmallmatrix}1 & 0 & 10^{-6}\\ 0 & 1 & 10^{-6}\end{psmallmatrix}$, the third feature's column norm is negligible yet its $D_i$ is exactly $1.0$, because numerator and dominant denominator term scale identically with column magnitude. \textbf{A genuinely dropped feature and a substantially represented but relatively diffuse feature can report identical $D_i$}; distinguishing them requires checking $\lVert W_i \rVert^2$ directly. This check is built into the classification pipeline from the start, and Section~\ref{sec:diffuse} is the payoff.

% ==================================================================
\section{The Superposition Phase Diagram}
\label{sec:phase}
% ==================================================================

\subsection{Methodology}
Fixed $n_{\mathrm{hidden}} = 20$; overcompleteness ratio $n_{\mathrm{features}}/n_{\mathrm{hidden}} \in \{1,2,4,8\}$; sparsity $s \in \{0,\allowbreak 0.3,\allowbreak 0.6,\allowbreak 0.8,\allowbreak 0.9,\allowbreak 0.95,\allowbreak 0.99\}$; uniform importance throughout; \textbf{3 independent seeds per cell}, all numbers reported as mean $\pm$ standard deviation. Each feature is classified by final $D_i$: \emph{dedicated} ($D_i > 0.9$), \emph{superposed} ($0.1 \le D_i \le 0.9$), or a candidate ``not represented'' case ($D_i < 0.1$), which---per Section~\ref{sec:featdim}---is further split by $\lVert W_i \rVert^2$ into \emph{truly dropped} ($\lVert W_i \rVert^2 < 0.1$) versus \emph{diffusely shared} ($\lVert W_i \rVert^2 \ge 0.1$).

\subsection{A convergence artifact, caught and documented}
\label{sec:convergence-bug}
The first full sweep produced an impossible result at ratio $=1$ ($n_{\mathrm{features}} = n_{\mathrm{hidden}} = 20$, where capacity suffices for a fully orthogonal solution and the correct converged answer is 100\% dedicated at every sparsity): at $s = 0.95$ and $0.99$, roughly 8\% and 82\% of features respectively classified as superposed. There is no capacity pressure forcing superposition in this configuration, so the result was investigated rather than reported. Re-running the suspicious cells at 8{,}000 training steps instead of the sweep default of 1{,}500 fully resolved the discrepancy: dedicated fraction returns to exactly $1.00 \pm 0.00$ with loss $\approx 0$ at every sparsity. The mechanism: at high sparsity a given feature appears in few training batches, so its gradient signal arrives rarely---convergence is \emph{slower}, not \emph{harder}---and a flat step budget tuned for low-sparsity cells silently under-trains the rest. The fix allocates the larger budget specifically to the affected cells. We report this in full because ``looked wrong, checked why, fixed it'' is the epistemic standard every other number in this paper is held to; under-training masquerading as superposition is a failure mode any replication of~\cite{elhage2022toy} at high sparsity should check for.

\subsection{Results}
\textbf{Ratio $=1$ (capacity-sufficient control):} dedicated fraction $1.00 \pm 0.00$, superposed $0.00 \pm 0.00$, loss $\approx 0$, at all seven sparsity levels. With no capacity pressure the model never chooses superposition; this row is as much a pipeline sanity check as a result.

\begin{table}[t]
\centering
\caption{Dedicated-feature fraction (mean $\pm$ std over 3 seeds) under genuine overcompleteness. More overcompleteness requires \emph{less} sparsity before superposition dominates. $^{*}$At ratio $8$, $s=0.99$, the $0\%$ dedicated coincides with $0\%$ superposed: a third regime, not an omission (Section~\ref{sec:diffuse}).}
\label{tab:phase}
\begin{tabular}{lccc}
\toprule
Sparsity $s$ & Ratio $=2$ & Ratio $=4$ & Ratio $=8$ \\
\midrule
0.00 & $1.00 \pm 0.00$ & $1.00 \pm 0.00$ & $1.00 \pm 0.00$ \\
0.30 & $0.26 \pm 0.01$ & $1.00 \pm 0.00$ & $1.00 \pm 0.00$ \\
0.60 & $0.02 \pm 0.02$ & $0.51 \pm 0.02$ & $0.47 \pm 0.07$ \\
0.80 & $0.00 \pm 0.00$ & $0.00 \pm 0.00$ & $0.00 \pm 0.00$ \\
0.90 & $0.00 \pm 0.00$ & $0.00 \pm 0.00$ & $0.00 \pm 0.00$ \\
0.95 & $0.00 \pm 0.00$ & $0.00 \pm 0.00$ & $0.00 \pm 0.00$ \\
0.99 & $0.00 \pm 0.00$ & $0.00 \pm 0.00$ & $0.00 \pm 0.00^{*}$ \\
\bottomrule
\end{tabular}
\end{table}

\begin{figure}[t]
\centering
\begin{subfigure}[b]{0.48\textwidth}
  \includegraphics[width=\linewidth]{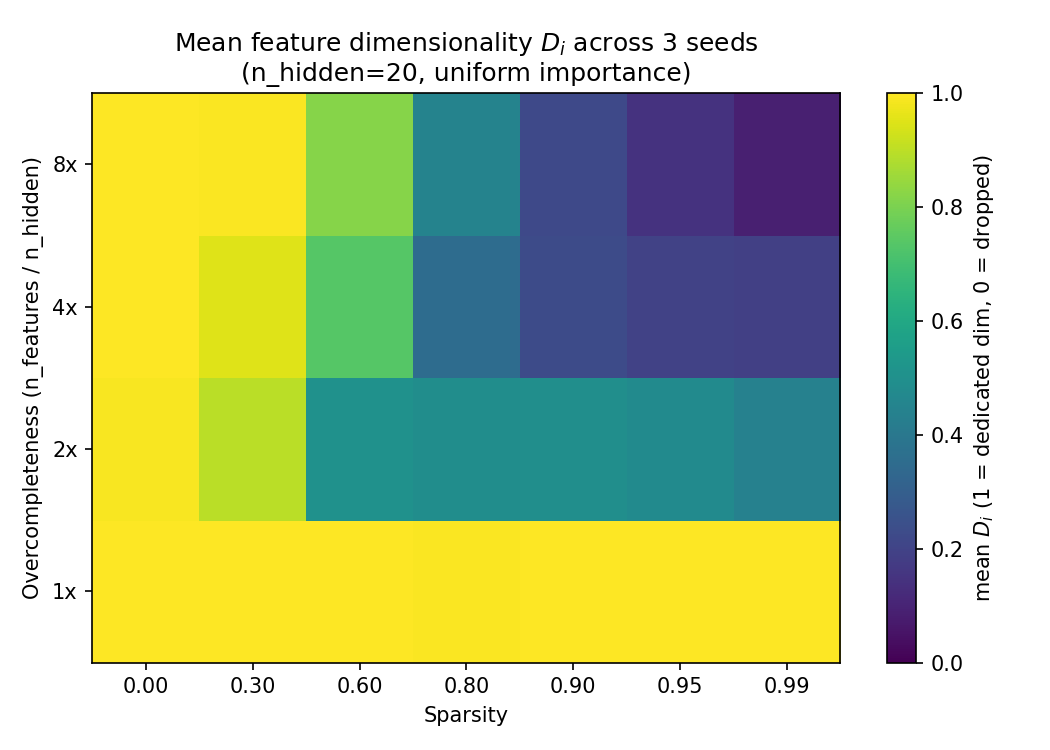}
  \caption{Mean $D_i$ over the sparsity $\times$ overcompleteness grid.}
\end{subfigure}\hfill
\begin{subfigure}[b]{0.48\textwidth}
  \includegraphics[width=\linewidth]{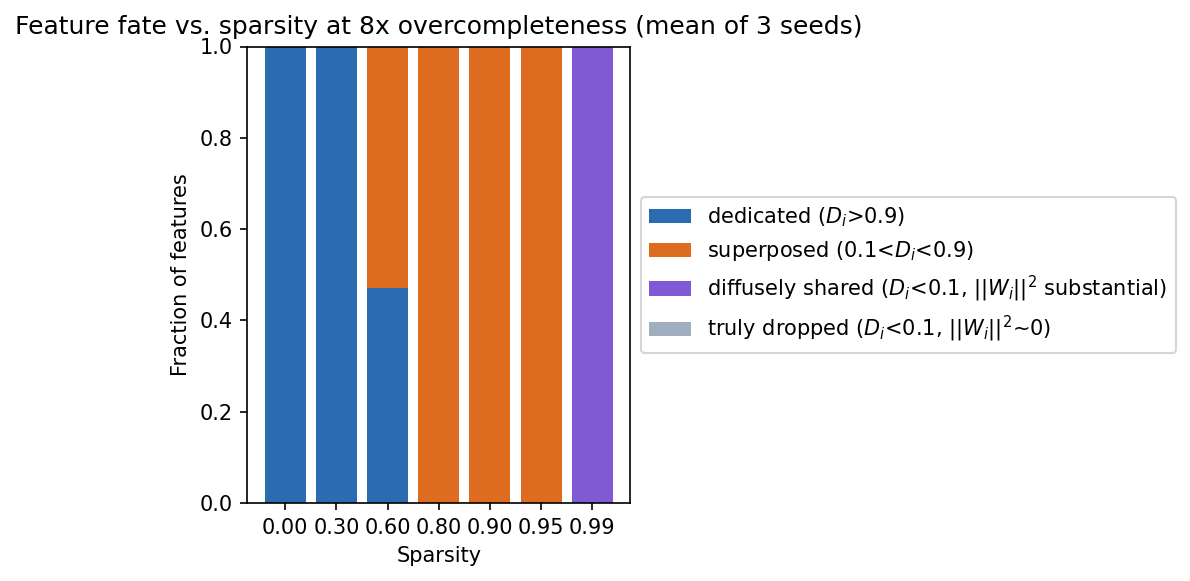}
  \caption{Regime fractions vs.\ sparsity at $8\times$ overcompleteness.}
\end{subfigure}\\[4pt]
\begin{subfigure}[b]{0.62\textwidth}
  \includegraphics[width=\linewidth]{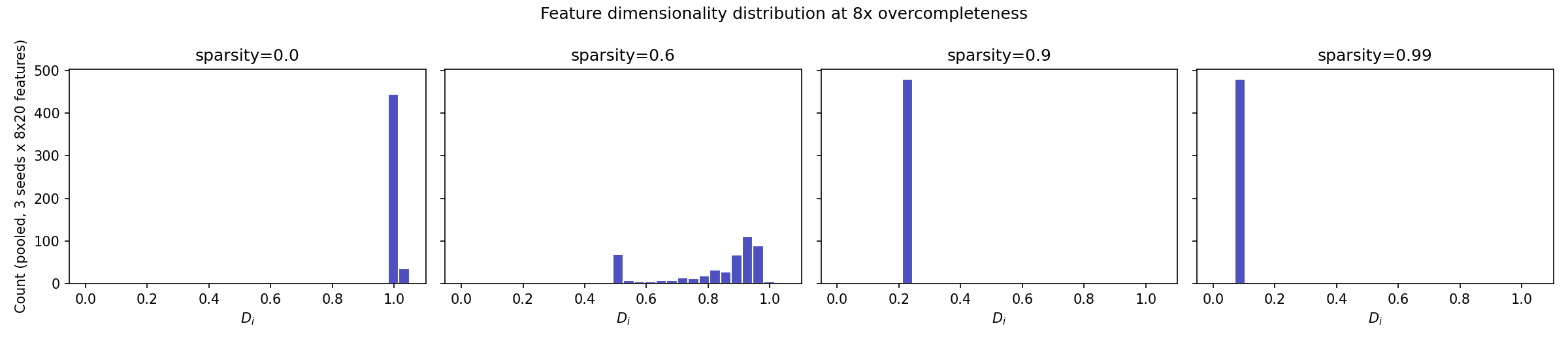}
  \caption{Pooled $D_i$ distributions at $8\times$ overcompleteness: a single peak at $D_i = 1$ ($s = 0$), a genuine spread ($s = 0.6$--$0.9$), and the tight diffuse cluster near $D_i = 0.08$ ($s = 0.99$).}
\end{subfigure}
\caption{The superposition phase diagram under uniform importance, 3 seeds per cell.}
\label{fig:phase}
\end{figure}

The qualitative pattern (Table~\ref{tab:phase}, Figure~\ref{fig:phase}) matches the theory directly. At $2\times$ overcompleteness the transition is well underway by $s = 0.3$ (74\% superposed); at $4\times$ and $8\times$ it occurs later (near $s = 0.6$) and at a similar critical sparsity for both---consistent with the picture that once superposition is worth adopting at all, further overcompleteness changes \emph{how much} is packed rather than \emph{whether} packing begins.

\subsection{An unplanned third regime: diffuse sharing}
\label{sec:diffuse}
The most extreme cell---ratio $8$, $s = 0.99$ (160 features into 20 dimensions, 99\% of inputs zeroed per sample)---reports $D_i \approx 0.08$ for every feature, below the $0.1$ ``candidate not-represented'' threshold; the raw classification would call all 160 features dropped. Checking $\lVert W_i \rVert^2$ directly tells the opposite story: mean $\lVert W_i \rVert^2 \approx 1.40$, \emph{larger} than a cleanly represented orthogonal feature's norm ($\lVert W_i \rVert^2 = 1$ at the ratio-1 optimum). Every feature is substantially represented, in a highly symmetric configuration where each overlaps slightly with very many others rather than owning a dimension (dedicated) or sharing exactly one with exactly one partner (antipodal). Reconstruction loss at this cell is low ($0.136 \pm 0.009$): the model succeeds via a third geometric strategy that the clean dedicated/antipodal/dropped trichotomy does not name. Whether this many-way diffuse packing is a documented phase or an under-described corner of the diagram is left open here; it is invisible unless $\lVert W_i \rVert^2$ is checked alongside $D_i$, which is why that check exists in the pipeline (Section~\ref{sec:featdim}).

% ==================================================================
\section{Dictionary Learning: $L_1$ versus TopK}
\label{sec:sae}
% ==================================================================

\subsection{Formulations}
The $L_1$ SAE is the classical convex relaxation of the NP-hard $\Ltop$ objective:
\begin{equation}
\mathcal{L}_{L_1} = \lVert h - \hat{h} \rVert_2^2 + \lambda \lVert f \rVert_1 ,
\end{equation}
which inherits Lasso shrinkage~\cite{tibshirani1996}: every active coefficient is pushed toward zero regardless of its true optimal value, so reported activation strengths systematically under-estimate feature presence. The TopK SAE~\cite{gao2024scaling} removes the penalty and enforces sparsity structurally---the encoder keeps the $k$ largest pre-activations (ReLU-rectified) and zeroes the rest, so $k$ directly \emph{is} the $\Ltop$:
\par\vspace{6pt}
\noindent\begin{minipage}{\linewidth}
\begin{lstlisting}
def encode(self, h):
    pre_act = (h - self.b_dec) @ self.W_enc + self.b_enc
    topk_vals, topk_idx = torch.topk(pre_act, k=self.config.k, dim=-1)
    topk_vals = torch.relu(topk_vals)
    f = torch.zeros_like(pre_act)
    f.scatter_(-1, topk_idx, topk_vals)
    return f
# loss = ((h - h_hat)**2).sum(-1).mean()   # no L1 term at all
\end{lstlisting}
\end{minipage}
\vspace{6pt}\par
With nothing to game by shrinking magnitudes, the only pressure on the $k$ surviving activations is reconstruction. TopK introduces its own failure mode---atoms that stop winning the selection stop receiving gradient and ``die''---mitigated here by a simplified version of the auxiliary loss of~\cite{gao2024scaling} (atoms silent within a rolling window receive a secondary residual-reconstruction objective). This is adequate to demonstrate the mechanism at toy scale, not a production implementation.

\subsection{Pareto comparison}
Both variants are trained on hidden activations of the same 32$\to$8, $s = 0.95$, decayed-importance toy model (22 well-represented features, 10 dropped), with $16\times$-overcomplete dictionaries (128 atoms), \textbf{3 seeds per configuration}. \emph{Recovery precision} counts well-represented ground-truth features whose best decoder-atom match has unsigned cosine $\ge 0.90$.

\begin{table}[t]
\centering
\caption{$L_1$ vs.\ TopK on identical superposed activations (mean $\pm$ std, 3 seeds).}
\label{tab:pareto}
\begin{tabular}{llccc}
\toprule
Method & Hyperparam. & $\Ltop$ & Recon.\ loss & Recovery precision \\
\midrule
$L_1$ & $\lambda = 0.003$ & $14.68 \pm 0.58$ & $0.00309 \pm 0.00039$ & $0.15 \pm 0.06$ \\
$L_1$ & $\lambda = 0.01$  & $10.69 \pm 0.58$ & $0.00334 \pm 0.00021$ & $0.23 \pm 0.10$ \\
$L_1$ & $\lambda = 0.03$  & $7.90 \pm 0.73$  & $0.00515 \pm 0.00029$ & $0.27 \pm 0.07$ \\
$L_1$ & $\lambda = 0.1$   & $4.42 \pm 0.32$  & $0.01427 \pm 0.00074$ & $0.77 \pm 0.11$ \\
\midrule
TopK & $k = 13$ & $12.92 \pm 0.07$ & $0.00031 \pm 0.00010$ & $0.11 \pm 0.06$ \\
TopK & $k = 9$  & $8.99 \pm 0.00$  & $0.00060 \pm 0.00009$ & $0.23 \pm 0.07$ \\
TopK & $k = 6$  & $6.00 \pm 0.00$  & $0.00166 \pm 0.00104$ & $0.53 \pm 0.13$ \\
TopK & $k = 4$  & $4.00 \pm 0.00$  & $0.01064 \pm 0.00210$ & $\bm{1.00 \pm 0.00}$ \\
\bottomrule
\end{tabular}
\end{table}

Two matched-$\Ltop$ comparisons carry the result (Table~\ref{tab:pareto}, Figure~\ref{fig:pareto}). At $\Ltop \approx 4$, TopK reaches \emph{perfect} recovery precision on the well-represented set ($1.00 \pm 0.00$ across all seeds---every one of the 22 represented features recovered, every seed) against $L_1$'s $0.77 \pm 0.11$. At $\Ltop \approx 9$--$13$, TopK's reconstruction loss is roughly an order of magnitude lower at comparable sparsity, while precision degrades on a similar curve for both methods as $\Ltop$ grows (more simultaneously active atoms means more cross-feature interference per match). TopK dominates the Pareto front at every $\Ltop$ tested---the qualitative claim of~\cite{gao2024scaling}, reproduced from scratch on an unrelated synthetic model.

\subsection{Direct evidence of shrinkage}
\label{sec:shrinkage}
Dominance alone does not establish \emph{why}; it is consistent with TopK being generically better for unrelated reasons. Gao et al.\ propose a falsifiable mechanism (shrinkage) with a specific test, reproduced exactly here: freeze which atoms each SAE selected for a held-out batch (its support), then re-optimize \emph{only the magnitudes} of those atoms---projected gradient descent, non-negativity constraint, decoder frozen---to directly minimize reconstruction error. If the SAE's reported activations were already optimal given its own support, refinement cannot improve anything.

\begin{table}[t]
\centering
\caption{Support-frozen magnitude refinement at matched sparsity ($\Ltop \approx 4$; $k{=}4$ TopK vs.\ $\lambda{=}0.1$ $L_1$; held-out batch of 5{,}000 activations).}
\label{tab:shrinkage}
\begin{tabular}{lccc}
\toprule
Method & $\Ltop$ & MSE improvement & Mean active-magnitude change \\
\midrule
$L_1$  & 4.23 & \textbf{91.2\%} & $\bm{+22.5\%}$ \\
TopK   & 4.00 & 64.8\%          & $-0.05\%$ (indistinguishable from zero) \\
\bottomrule
\end{tabular}
\end{table}

\begin{figure}[t]
\centering
\begin{subfigure}[b]{0.55\textwidth}
  \includegraphics[width=\linewidth]{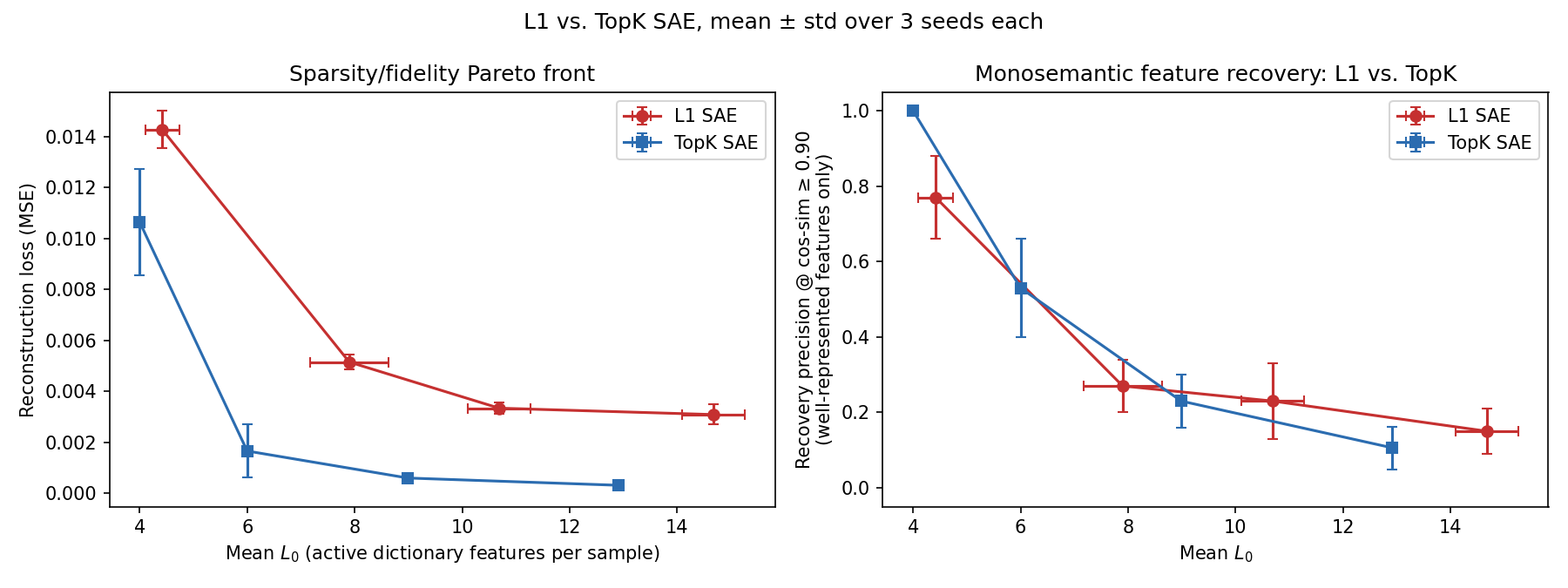}
  \caption{$\Ltop$ vs.\ reconstruction loss and vs.\ recovery precision (error bars over 3 seeds).}
  \label{fig:pareto}
\end{subfigure}\hfill
\begin{subfigure}[b]{0.42\textwidth}
  \includegraphics[width=\linewidth]{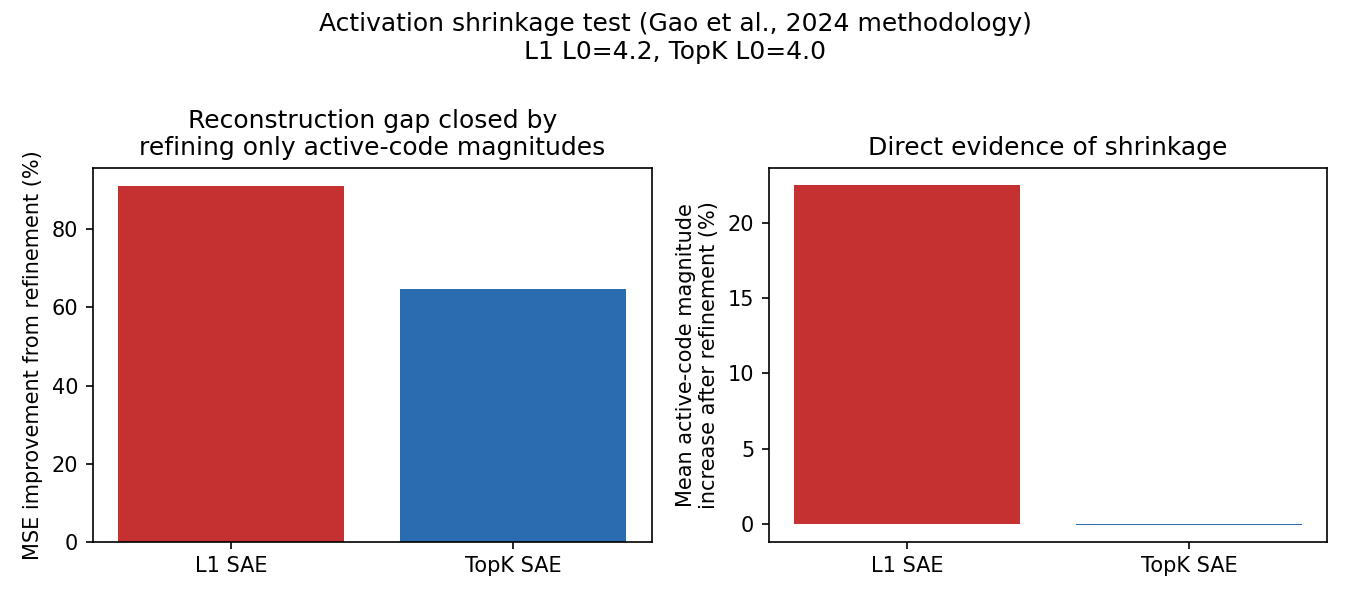}
  \caption{Refinement MSE improvement and magnitude change, $L_1$ vs.\ TopK.}
  \label{fig:shrinkage}
\end{subfigure}
\caption{The $L_1$/TopK comparison and the shrinkage mechanism test.}
\end{figure}

Refining the $L_1$ SAE's activations closes 91\% of its reconstruction gap by pushing magnitudes up nearly a quarter on average---direct quantitative evidence that the penalty suppresses true feature strengths, exactly as the Lasso analysis predicts (Table~\ref{tab:shrinkage}, Figure~\ref{fig:shrinkage}). TopK refinement still closes some gap (refinement is better-or-equal by construction; the originals came from one gradient trajectory, not an exact solve) but its mean magnitude does not move: there is no systematic direction to correct because no penalty pushed in one. Practical consequence: an $L_1$ dictionary's activation \emph{strengths} are systematically biased low---relevant to any downstream use ranking features by magnitude, thresholding anomaly signals, or comparing strength across inputs---while TopK trades that bias for a rigid per-sample budget and the dead-atom problem.

% ==================================================================
\section{From Correlation to Causation}
\label{sec:causal}
% ==================================================================

Recovery precision of $1.00$ for TopK $k{=}4$ is still a correlational claim: a learned decoder direction \emph{points at} the right ground-truth feature. It is not a causal claim: that \emph{intervening} on the direction changes behavior the way the label implies. A decoder atom can sit at the geometrically correct location without the \emph{encoder} ever activating it when the feature is genuinely present---two failure modes hiding behind one cosine number. Closing the gap requires an intervention, not another measurement of the same representation.

\subsection{Ablation and steering}
For every well-represented ground-truth feature $i$, matched by cosine similarity to dictionary atom $j$, two interventions are propagated through the toy model's actual output stage $\hat{x} = \mathrm{ReLU}(\hat{h}W + b)$, never a proxy:
\begin{itemize}[leftmargin=1.6em,itemsep=1pt]
  \item \textbf{Ablation.} Present inputs where \emph{only} feature $i$ is active; encode; zero atom $j$'s code entry; decode; compare the targeted drop in feature $i$'s reconstructed output against the mean absolute movement of every other feature. Their ratio is the \emph{ablation specificity}: a causally faithful match hits its target hard and everything else barely at all; a ratio near 1 is indistinguishable from generic collateral damage.
  \item \textbf{Steering.} The reverse direction: inputs where feature $i$ is \emph{off}; force atom $j$ to a fixed value with the correct match \emph{sign}; measure the targeted rise.
\end{itemize}

\par\vspace{6pt}
\noindent\begin{minipage}{\linewidth}
\begin{lstlisting}
f_ablated = f.clone()
f_ablated[:, atom_idx] = 0.0
x_hat_ablated = torch.relu(sae.decode(f_ablated) @ toy_model.W + toy_model.b)

targeted_drop     = (x_hat_base[:, feat_idx] - x_hat_ablated[:, feat_idx]).mean()
off_target_effect = (x_hat_base - x_hat_ablated).abs()[:, other_features].mean()
specificity_ratio = targeted_drop / off_target_effect
\end{lstlisting}
\end{minipage}
\vspace{6pt}\par

\paragraph{A sign-convention bug, and what it demonstrates.}
Cosine matching takes absolute values---correctly, since a feature and its exact negation are equally good \emph{correlational} matches---but an intervention is direction-sensitive. The first steering implementation silently assumed positive alignment and always pushed the matched atom toward $+1$. The test suite caught it immediately: on a sanity-check model whose true match was \emph{anti}-aligned (dominant weight $-0.82$), steering ``on'' in the wrong direction pushed the reconstruction into a region the output ReLU clips, and the measured effect read exactly $0.0$ instead of positive. The fix recovers the signed match explicitly (\texttt{best\_match\_indices\_and\_signs}) and makes the sign a required argument of \texttt{steering\_effect}, so the bug cannot silently reappear through the same code path. Section~\ref{sec:instrument} promotes this from a patch to an interface contract; Section~\ref{sec:anatomy} shows the structure that fixing it revealed.

\subsection{Does a good correlational match predict a good causal match?}
Two TopK SAEs from opposite ends of the Pareto front are audited: $k = 4$ (recovery precision $1.00$; ``good'') and $k = 13$ (precision $0.11$; deliberately degraded, ``bad''). For each of the 22 well-represented features, both interventions run against each SAE's matched atom.

\begin{table}[t]
\centering
\caption{Causal specificity of correlationally matched features (original multi-threaded runs; see Section~\ref{sec:reaudit} for the deterministic re-audit).}
\label{tab:causal-v1}
\small
\setlength{\tabcolsep}{5pt}
\begin{tabular}{lcccc}
\toprule
SAE & Median abl.\ specificity & Min & Max & Median steer.\ specificity \\
\midrule
TopK $k{=}4$ (precision 1.00)  & \textbf{168.8} & $0.0$    & $2608.6$ & $149.9$ \\
TopK $k{=}13$ (precision 0.11) & \textbf{0.0}   & $-902.5$ & $163.0$  & $21.4$ \\
\bottomrule
\end{tabular}
\end{table}

\begin{figure}[t]
\centering
\includegraphics[width=0.9\textwidth]{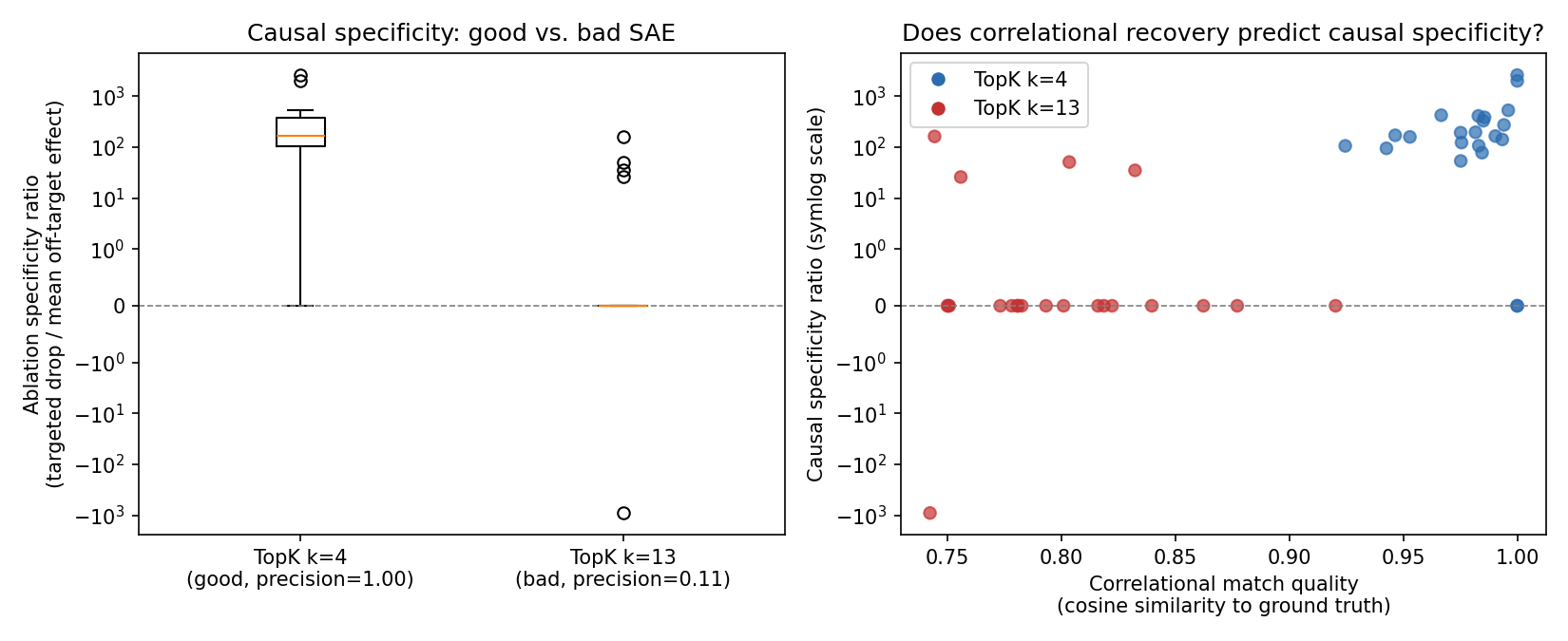}
\caption{Left: ablation specificity by SAE (symlog scale; some values are exactly zero or negative). Right: cosine similarity vs.\ specificity, both SAEs pooled ($r = 0.657$).}
\label{fig:causal-v1}
\end{figure}

The good SAE's median ablation specificity is 168.8---ablating the matched atom moves the target roughly 169 times more than everything else---against the bad SAE's median of \emph{exactly zero} (Table~\ref{tab:causal-v1}, Figure~\ref{fig:causal-v1}). Pooled correlation between cosine similarity and log-specificity: $r = 0.657$---a real positive relationship (better correlational matches are better causal matches on average), far from $r = 1$, and the exceptions are the informative part.

\subsection{Why the degraded SAE's median is exactly zero}
\label{sec:inert-v1}
Seventeen of the bad SAE's 22 matched pairs report ablation specificity $= 0.0$ \emph{exactly}. Tracking \firedfrac{} (the fraction of feature-present samples on which the matched atom's code is nonzero) alongside the ratio gives an unambiguous cause: in all seventeen cases the matched atom \emph{never fires} when the feature is presented in isolation ($\firedfrac = 0.0$ over 500 samples). Ablating an already-zero code entry cannot change anything; the ratio is not small but undefined-and-reported-as-zero, for a different reason than a weak effect. One of the seventeen (feature 20) has cosine similarity $0.92$---above the $\ge 0.90$ recovery bar used throughout. \textbf{By the field's correlational metric, feature 20 is a recovery success; by the causal test, the atom credited with recovering it never once activates for it.} This is 17 of 22 pairs (77\%), not a dredged-up edge case. The good SAE exhibits the same mode at a lower rate: 2 of 22 matches (9\%), both at cosine $> 0.9999$. Even the best configuration tested is not immune to geometrically perfect, causally inert matches.

\subsection{Mechanism: decoder geometry versus encoder selection}
Cosine matching---here and in the literature this reproduces---compares target directions against \emph{decoder} atoms $W_{\mathrm{dec}}$, the vectors used to reconstruct activations from a code. Whether an atom \emph{fires} is governed by the \emph{encoder} weights and the TopK/ReLU selection, not by decoder geometry. An SAE under pressure to reconstruct well can learn a decoder atom at the geometrically correct location---useful for the reconstruction objective it is actually trained on---while the paired encoder entry essentially never wins the TopK competition for that feature, because something else usually wins first. Decoder-geometry recovery and encoder-activation recovery are two different empirical claims that the correlational metric cannot, by construction, distinguish. Only an intervention can.

% ==================================================================
\section{\texttt{sae-causal-audit}: The Methodology as an Instrument}
\label{sec:instrument}
% ==================================================================

A finding locked inside its own experiment scripts is a claim, not a tool: anyone wanting \emph{their} inert rate would have to separate methodology from experiment and re-implement the former. The methodology is therefore packaged as \texttt{sae-causal-audit}, an instrument that must survive inputs its author never anticipated. That constraint forced the design decisions below, surfaced two further bugs (Section~\ref{sec:repro}), and refined the original finding itself (Section~\ref{sec:anatomy}).

\subsection{Structural typing: what ``any SAE'' means}
The audit accepts anything satisfying a three-member runtime-checkable protocol---no inheritance, no registration, no adapter:
\par\vspace{6pt}
\noindent\begin{minipage}{\linewidth}
\begin{lstlisting}
@runtime_checkable
class SparseAutoencoder(Protocol):
    W_dec: torch.Tensor                                  # (d_sae, d_in)
    def encode(self, h: torch.Tensor) -> torch.Tensor: ...
    def decode(self, f: torch.Tensor) -> torch.Tensor: ...
\end{lstlisting}
\end{minipage}
\vspace{6pt}\par
\texttt{sae\_lens.SAE}---the interface behind most published open-weight suites---satisfies it out of the box. The division of labor restates the central finding structurally: \textbf{matching uses \texttt{W\_dec}} (the correlational step \emph{should} use decoder geometry; that is what the field's metric measures and what is being audited), while \textbf{every causal metric routes through \texttt{encode}/\texttt{decode}}, so encoder-side selection---the thing decoder geometry cannot certify---is what is actually measured. Ground truth enters through a second protocol (\texttt{FeatureProbe}: \texttt{activations\_with\_feature}, \texttt{activations\_without\_feature}), the seam that lets one pipeline serve two regimes: synthetic isolation inputs (exact ground truth) in the toy regime; positive/negative prompt sets captured at a hook point (a stated, weaker substitution) in the real-model regime of Section~\ref{sec:real}.

\subsection{Three measurements per matched pair}
In increasing order of causal strength, all propagated through \texttt{encode}/\texttt{decode} and a caller-supplied behavioral readout:
\begin{enumerate}[leftmargin=1.8em,itemsep=1pt]
  \item \textbf{Fired fraction} (the cheap screen that alone explained all seventeen inert cases of Section~\ref{sec:inert-v1}):
  \begin{equation}
  \mathrm{fired\_frac}_{i,j} = \frac{1}{N}\sum_{n=1}^{N} \mathbb{1}\!\left[\,|f_j(h_n)| > \varepsilon\,\right],\qquad h_n \sim \text{feature $i$ present}.
  \end{equation}
  If the encoder never activates atom $j$ across hundreds of feature-present samples, no downstream claim about $j$ ``representing'' $i$ can be causal, whatever the cosine says. Cost: one batched \texttt{encode}.
  \item \textbf{Ablation specificity} (read direction), as in Section~\ref{sec:causal}.
  \item \textbf{Sign-correct steering specificity} (write direction), with the match sign a required argument.
\end{enumerate}

\subsection{Interface design as bug prevention}
The sign bug of Section~\ref{sec:causal} is made unrepresentable rather than merely fixed. The matcher returns both facts as separate fields instead of collapsing them into one lossy number:
\par\vspace{6pt}
\noindent\begin{minipage}{\linewidth}
\begin{lstlisting}
@dataclass(frozen=True, slots=True)
class MatchResult:
    feature_idx: int
    atom_idx: int
    cosine: float   # unsigned, in [0, 1] -- recovery semantics
    sign: float     # +/-1.0             -- intervention semantics
\end{lstlisting}
\end{minipage}
\vspace{6pt}\par
and \texttt{steering\_effect(...)} has no default sign to silently assume. The original failure is a permanent regression test: an anti-aligned atom steered with the wrong sign must read exactly $0.0$ through a ReLU downstream, and with the correct sign must read positive.

Two further semantics rules are load-bearing. \textbf{Zeros carry their cause}: specificity $0.0$ can mean ``weak effect'' or ``the code was already zero,'' with opposite implications; every causal result therefore ships with its \firedfrac, and the \texttt{causally\_inert} flag is \emph{defined} as $\firedfrac = 0.0$, never inferred from the ratio. \textbf{Infinity is legal where it means something and rejected where it does not}: a perfectly surgical intervention (nonzero target, exactly zero collateral) is legitimately infinite specificity; the serializer encodes it safely (strict JSON has no \texttt{Infinity} literal), median-based statistics accept it, and mean-based statistics refuse it loudly.

\subsection{Test suite as instrument calibration}
The suite (37 tests) works in three layers: \emph{hand-derivable unit cases} (identity SAEs and two-atom constructions where every expected value is computable on paper: a perfectly surgical ablation returns exactly $\infty$; a structurally dead atom returns exactly $0.0$ with $\firedfrac = 0.0$; equal mixing into one other dimension returns specificity exactly $1.0$---the same standard applied to $D_i$, where the antipodal pair must come out at exactly $0.5$); \emph{property-based tests} (Hypothesis) over randomized inputs (cosine stays in $[0,1]$; the reported sign agrees with the true signed cosine of the chosen pair; dead atoms never win a match through a $0/0$; bootstrap intervals are ordered and bracketed by the sample range); and an \emph{end-to-end integration layer} that trains the real toy setting---real optimizer, real TopK competition---and asserts the audit recovers the known qualitative answers, with any shape mismatch failing loudly rather than producing numbers on misaligned tensors.

% ==================================================================
\section{Reproducibility as a Stack of Claims}
\label{sec:repro}
% ==================================================================

The package makes a specific, checkable promise: two runs producing identical science produce byte-identical result files, letting CI regenerate every result and compare SHA-256 hashes---regression detection on the \emph{results}, not just the code. Actually executing that promise, rather than asserting it, surfaced two bugs (the third and fourth caught across this line of work) and forced the promise itself to be restated.

\paragraph{A serialization leak.}
\begin{sloppypar}
The first attempted proof failed: two back-to-back from-scratch reproductions---same seeds, single-threaded, \texttt{torch.use\_deterministic\_\allowbreak algorithms(True)}---produced different hashes with identical scientific numbers. The report dataclass carried a \texttt{runtime\_seconds} field; wall-clock timing was being serialized into the hashed JSON. The fix is a documented \texttt{VOLATILE\_FIELDS} exclusion set: timing remains available in memory and in the human-readable render and never enters the hashed artifact. The gate is tested the only way a gate honestly can be: reproduce twice and require identical hashes, then deliberately corrupt one number and require the gate to fail, naming the file with expected and observed hashes.
\end{sloppypar}

\paragraph{Three layers of cross-environment nondeterminism.}
The same gate, run on CI, produced different hashes on different runs of the same commit---same seeds, same code, same pinned dependencies. Unwinding it went three layers deep:
\begin{enumerate}[leftmargin=1.8em,itemsep=2pt]
  \item \textbf{The optimizer was multi-threaded behind the scenes.} \texttt{torch.set\_num\_threads(1)} pins intra-op parallelism, but Adam defaults to \texttt{foreach=True}---a fused multi-tensor path whose floating-point accumulation order differs from the naive loop's---and inter-op parallelism had never been pinned. Float addition is not associative; a different reduction order is a different last bit. Setting \texttt{foreach=False}, pinning interop threads to one, and adding defensive environment variables (\texttt{MKL\_CBWR}, BLAS thread counts, \texttt{PYTHONHASHSEED}) restored single-environment determinism.
  \item \textbf{The degraded SAE is a noise amplifier---a finding, not an inconvenience.} A last-bit difference should change a hash and nothing else; in the $k{=}13$ SAE it changed \emph{integer counts}. One feature's best-match cosine sits close enough to the $0.90$ bar, and one atom's pre-activation close enough to the TopK cut, that last-bit noise flips them across. The good SAE's census never moved. The same TopK competition that produces causally inert matches also produces numerical boundary instability---one mechanism, two symptoms, at two different levels of the stack (Section~\ref{sec:anatomy} closes this loop). The repository encodes an explicit $\pm 1$ tolerance band on that SAE's counts, and only that SAE's.
  \item \textbf{``The same torch version'' is not the same program.} With CI-side determinism proven, hashes still refused to match a baseline generated on Windows: the \texttt{torch==2.13.0+cpu} wheels for Windows and Linux are different binaries (MSVC vs.\ GCC~13.3; MKL~2026.1 vs.\ 2024.2; NNPACK off vs.\ on). Different compiled code rounds differently in the last bit---deterministically, permanently. No environment variable bridges it. \textbf{Byte-exact cross-platform reproducibility is not hard; it is unavailable, by construction.} The visible symptom was layer two in different clothes: the degraded SAE's census read one feature higher on Windows while every continuous metric agreed to four decimal places.
\end{enumerate}

\paragraph{The restructured promise.}
The repository now makes two guarantees with disjoint scopes. \textbf{Byte-exact} (SHA-256-identical result files) is guaranteed within the pinned CI environment (\texttt{ubuntu-24.04}, \texttt{torch==2.13.0+cpu}) and verified on every push; the gate has produced identical hashes across three independent runs of one commit. \textbf{Semantic} (every number within $\mathrm{rtol} = 10^{-4}$; boundary-sensitive counts within $\pm 1$) is guaranteed on any platform. The byte-exact baseline is generated by the CI environment itself via a manually dispatched job---never regenerated locally---because a baseline produced on the wrong machine \emph{was} the entire third layer. The tolerance tier is tested like the hash tier: corrupt a count by two, or nudge a continuous metric by one percent, and the gate must fail; both do. The generalization we propose: \emph{reproducibility is not one claim but a stack of claims with different scopes, and an honest repository states which rung it guarantees where.} Most ``fully reproducible'' badges silently mean \emph{on my machine, probably}.

% ==================================================================
\section{Re-audit Under the Deterministic Pipeline}
\label{sec:reaudit}
% ==================================================================

The deterministic pipeline (single-threaded, \texttt{foreach=False}, deterministic algorithms enforced, fixed seeds; $\sim$3 minutes on CPU) retrains the original configuration and audits the same two TopK SAEs ($k{=}4$, $k{=}13$; 128-atom dictionaries). One environmental note deserves a full sentence: the fully deterministic pipeline converges to \emph{different weights} than the original multi-threaded runs, so the exact counts moved while the qualitative finding did not---precisely the non-transferability the original analysis warned about when it noted that a number measured for one checkpoint is not a portable constant. This is an accidental robustness check the claim passed. All numbers below come from the reference environment whose hashes CI verifies; where boundary sensitivity makes a count environment-dependent, it is stated at the number.

\begin{table}[t]
\centering
\caption{The causal census over the 22 well-represented features, cosine $\ge 0.90$ recovery bar, deterministic pipeline. Bracketed ranges are 95\% bootstrap confidence intervals. $^{\dagger}$Medians and intervals are computed over every recovered pair across all 32 features ($n = 24$ and $n = 21$ respectively), not only the well-represented 22 used for the census rows.}
\label{tab:census}
\small
\setlength{\tabcolsep}{4pt}
\begin{tabular}{lcc}
\toprule
 & TopK $k{=}4$ (good) & TopK $k{=}13$ (bad) \\
\midrule
Correlationally recovered, of 22 & \textbf{22 / 22} & 18 / 22 \\
Recovered but \textbf{causally inert} ($\firedfrac = 0$) & 2 (\textbf{9\%}) & 3 (\textbf{17\%}); 4 in $\pm 1$ band \\
Median ablation specificity$^{\dagger}$ & \textbf{133.8} \ [107.0--167.4] & 68.3 \ [\textbf{22.0}--105.9] \\
Median steering specificity$^{\dagger}$ & 37.7 \ [33.1--41.8] & 16.9 \ [14.2--21.3] \\
\bottomrule
\end{tabular}
\end{table}

\begin{figure}[t]
\centering
\begin{subfigure}[b]{0.48\textwidth}
  \includegraphics[width=\linewidth]{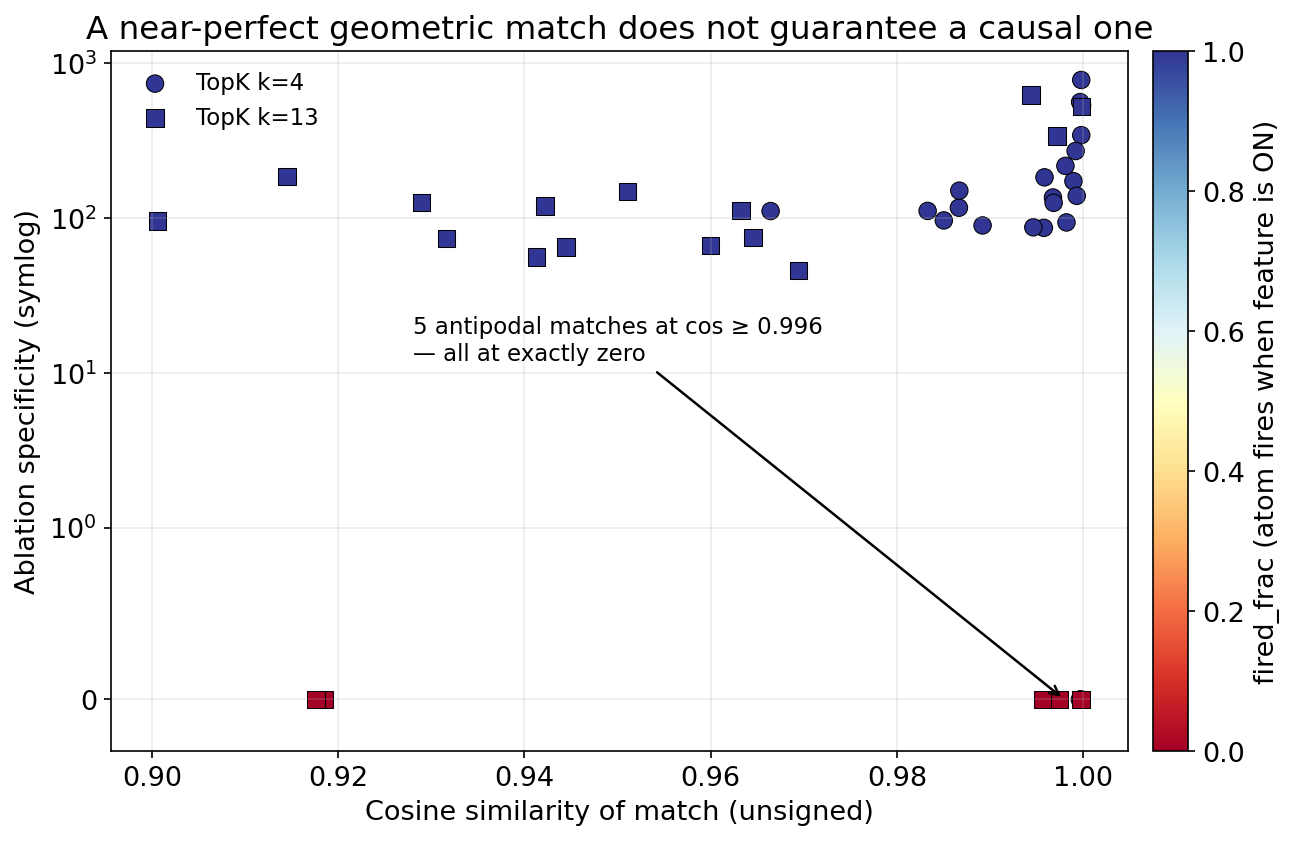}
  \caption{Cosine vs.\ ablation specificity (symlog), colored by \firedfrac; high-cosine points span from specificity in the hundreds to exactly zero.}
\end{subfigure}\hfill
\begin{subfigure}[b]{0.48\textwidth}
  \includegraphics[width=\linewidth]{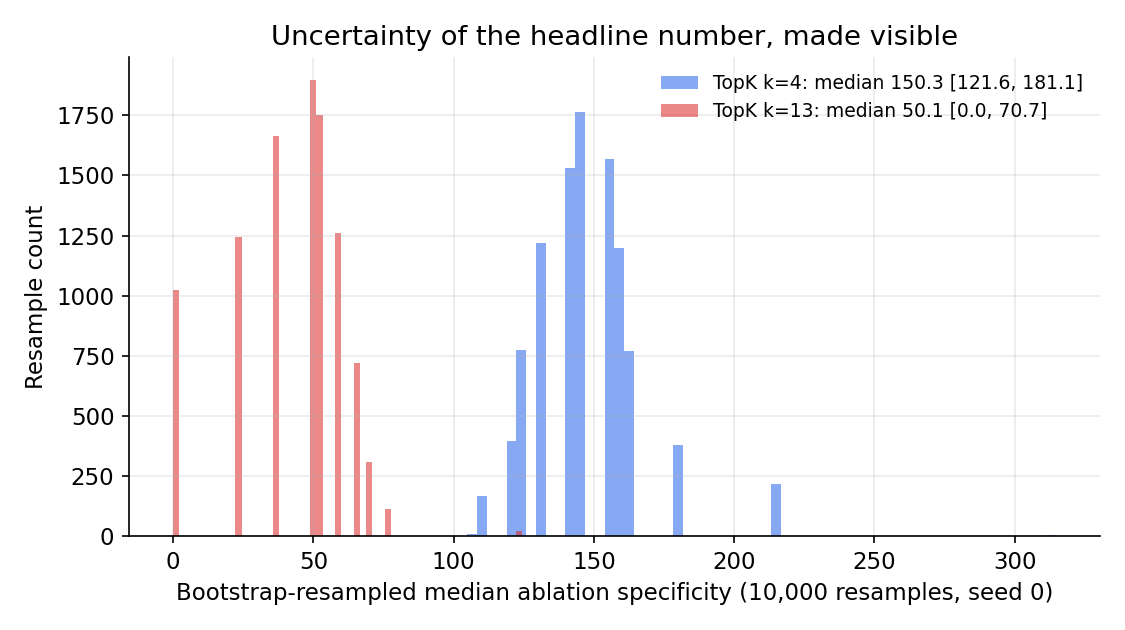}
  \caption{10{,}000 bootstrap-resampled medians per SAE: the good SAE's distribution sits tightly; the bad SAE's is wide, multimodal, with mass reaching toward zero.}
\end{subfigure}
\caption{The re-audit under the deterministic pipeline.}
\label{fig:reaudit}
\end{figure}

Three observations in Table~\ref{tab:census} (Figure~\ref{fig:reaudit}) deserve slow reading.

\textbf{First, the inert features are not marginal matches---in either SAE.} The bad SAE's inert cosines are $0.986$, $0.994$, and $0.9995$; the good SAE's two are $\bm{0.9997}$ and $\bm{0.9998}$. The \emph{good} SAE---whose recovered set is otherwise causally excellent---contains features whose geometric match is perfect to three decimal places and whose atoms do not fire once across 500 feature-present samples. The original experiment's most alarming single example was inert at cosine $0.92$; the instrument, on a fresh run, finds inert matches at effectively $1.000$, in both SAEs. The correlational metric is not merely noisy at the margin; it can be maximally confident and causally wrong, and SAE quality does not prevent it. Section~\ref{sec:anatomy} gives the clean geometric reason the good SAE, of all things, produces the most perfect inert matches.

\textbf{Second, the correlational metric is least trustworthy exactly where it is most needed.} Correlation between cosine and log ablation specificity, over each SAE's \emph{firing} pairs: $r = 0.859$ ($n = 21$) for the good SAE versus $r = 0.471$ ($n = 18$) for the bad one. The earlier pooled figure ($r = 0.657$) hid the decision-relevant shape: on a healthy dictionary cosine is a decent (never sufficient) proxy for causal quality; on a degraded dictionary---the case an audit exists to catch---the proxy itself degrades toward uninformative. Cosine similarity is a fair-weather instrument.

\textbf{Third, the degraded SAE's uncertainty structure is itself the finding.} The good SAE's 95\% bootstrap interval on median ablation specificity spans 107--167 around 134 (relative width $\approx 45\%$); the bad SAE's spans 22--106 around 68 (relative width $\bm{123\%}$), its lower bound within an order of magnitude of the specificity-1 collateral-damage line. A percentile bootstrap over an 18-pair recovered set containing three exact zeros produces resampled medians swinging across most of the observed range. Every summary number ships with a seeded, 10{,}000-resample interval~\cite{efron1993bootstrap}---the upgrade the original $n \approx 20$ point estimates explicitly owed.

The count carrying the $\pm 1$ annotation is itself data: on the reference Linux environment the bad SAE's census reads 18 recovered with 3 inert; on a Windows build of the same torch version it reads 19 with 4, because one non-antipodal feature (cosine $0.924$) sits close enough to both the recovery bar and the TopK selection boundary that different BLAS builds resolve it differently. That flip-prone feature is the competitive-inertness case of the next section---the boundary sensitivity of Section~\ref{sec:repro} and the causal taxonomy below are the same phenomenon seen from two altitudes.

% ==================================================================
\section{Anatomy of the Perfect Inert Matches}
\label{sec:anatomy}
% ==================================================================

\subsection{Antipodal pairs, end to end}
The good SAE's two inert features are the same geometric event twice. Feature 8 matches atom 78 at cosine $0.9997$, \emph{anti}-aligned (sign $-$), with $\firedfrac = 0.00$: ablation specificity exactly 0. But atom 78 is simultaneously the best match for \emph{feature 1}---same atom, positively aligned, firing on 100\% of feature-1 samples with ablation specificity 535. One atom, two ``recovered'' features. The other inert feature is the identical story at different indices: feature 5 matches atom 22 at cosine $0.9998$, anti-aligned, never fires, while feature 0 matches the same atom positively aligned, firing every time at specificity 347.

The toy model's weights explain everything: $\cos(W_1, W_8) = -0.9998$ and $\cos(W_0, W_5) = -0.9999$. Both are \textbf{antipodal pairs}---the exact $W_i \approx -W_j$ configuration the superposition analysis predicts (Section~\ref{sec:phase})---two well-represented features sharing one direction with opposite signs. The SAE, reasonably, learned \emph{one atom per shared axis}. The encoder's TopK-then-ReLU selection passes only positive pre-activations, so each atom fires for the positive side of its axis and never for the negative side. Decoder geometry for the negative-side feature is essentially perfect; encoder behavior for it is essentially nonexistent. This answers why the \emph{good} SAE produces the most perfect inert matches: \textbf{antipodal inertness is not a training failure; it is superposition geometry faithfully compressed.} A better SAE learns the shared axis \emph{more} precisely, pushing the anti-aligned cosine \emph{closer} to $1.000$. The failure is structural, and no amount of SAE quality removes it---only interface-level awareness of match sign does. The bad SAE replicates the pattern threefold (atoms 22, 21, 39 each serving pairs 0/5, 1/8, 4/2, each pair anti-parallel in ground truth to at least $-0.999$). Across both SAEs: \textbf{five antipodal pairs, accounting on the reference environment for every causally inert feature in both recovered sets.}

\subsection{A two-axis taxonomy of causal inertness}
\label{sec:taxonomy}
\textbf{By cause.} \emph{Structural inertness}---the antipodal mechanism---appears in good and bad SAEs alike, survives improved training, and is diagnosable from geometry alone: a shared atom matched with opposite signs is visible in the match table before any intervention runs. It is a hierarchy-free sibling of feature absorption~\cite{chanin2024absorption}: both produce latents that do not fire where their concept is present and both resist being trained away, but absorption is driven by the sparsity objective over hierarchical features, while structural inertness needs nothing beyond an antipodal pair and a positive-pass encoder. \emph{Competitive inertness}---an atom that keeps losing the TopK competition despite decent geometry, the dominant mechanism behind the original 77\% figure---appears only in the degraded SAE. Its fourth inert case, the non-antipodal feature at cosine $0.924$ that flips in and out of the census across BLAS builds, is this second kind, and the coincidence is not one: an atom that \emph{barely} loses the TopK competition is an atom whose selection is decided in the last bits of a float. Competitive inertness and numerical boundary sensitivity are one mechanism expressed at two levels. Operationally the two kinds are opposites: structural inertness is stable, predictable, and screenable from the dictionary alone; competitive inertness is unstable by nature and is what the \firedfrac{} screen exists to catch empirically.

\textbf{By direction.} The antipodal features are ablation-inert---one cannot remove what never activates---but they are \emph{not} steering-inert: forcing atom 78 on \emph{with the correct negative sign} raises feature 8's reconstructed output at steering specificity $\bm{310}$; the same intervention on feature 5's atom scores $203$; the bad SAE's three antipodal cases score $143$--$261$. These are among the highest steering specificities in the entire audit, attached to atoms whose ablation effect is exactly zero. The decoder direction is causally \emph{usable} in the write direction even though the encoder never uses it in the read direction. ``Causally inert'' therefore decomposes into two separable claims---\textbf{read-inert} (the encoder never selects the atom for the feature: ablation-based monitoring built on it is blind) and \textbf{write-inert} (interventions along the atom do not move the feature: steering built on it is impotent)---and all five antipodal pairs dissociate the two completely. A feature can be unmonitorable yet steerable through the same atom; in this setting that is not an edge case but the \emph{systematic signature} of antipodal superposition under a positive-pass encoder. This decomposition was invisible in the original experiment: without the signed-steering machinery working correctly, the write-direction test on anti-aligned matches returned zeros for the wrong reason. Fixing the instrument revealed structure the bug had been hiding.

Two design notes fall out of the same examples. The matcher is deliberately greedy (per-feature argmax) rather than a bipartite assignment, precisely so atom collisions---one atom best-matching two features---surface in the results instead of being optimized away; a collision is a finding about the dictionary (one atom serving an antipodal pair), not matching noise, and the audit's five collisions were the thread that unraveled everything above. And the \firedfrac{} screen would have flagged all five for the cost of one batched \texttt{encode}, before any intervention ran---at production scale, the difference between a cheap census and an expensive one.

% ==================================================================
\section{Toward Production SAEs: The Real-Model Harness}
\label{sec:real}
% ==================================================================

The package ships \texttt{scripts/audit\_real\_sae.py}: the identical pipeline pointed at published production SAEs---GPT-2 residual-stream SAEs and Gemma Scope~\cite{lieberum2024gemmascope}---via SAELens and TransformerLens. The real setting forces two substitutions, made explicit rather than smuggled in:
\begin{itemize}[leftmargin=1.6em,itemsep=1pt]
  \item \textbf{Ground truth $\to$ probe datasets.} ``Feature present'' is defined by positive/negative prompt sets per labeled concept, activations captured at the SAE's hook point on the final token; the concept direction needed for cosine matching is a difference-in-means linear-probe direction---explicitly the weakest link in the real regime, a proxy standing in for ground truth that does not exist, and precisely why the pipeline is calibrated first in the toy regime where the test suite can assert the right answers.
  \item \textbf{Toy readout $\to$ spliced logits.} The downstream readout decodes the (possibly intervened) code, splices the reconstruction back into the residual stream at the hook point, runs the remainder of the model, and reads mean logits over concept-diagnostic tokens; ablation specificity then asks whether zeroing the atom suppresses \emph{this} concept's tokens more than it moves the other audited concepts'.
\end{itemize}
GPT-2-small runs on a free-tier T4 (or CPU, slowly); Gemma-2-2b with Gemma Scope wants $\sim$16\,GB of GPU memory. The research question the harness exists to answer is the obvious promotion of everything above: \emph{of the features in a published production SAE that pass a standard correlational bar for a labeled concept, what fraction never fire when the concept is actually present---and among those that fire, how does read-specificity relate to write-specificity?} We do not know, and we specifically do not predict that toy-regime rates transfer---the probe-direction proxy alone guarantees the numbers are not comparable one-to-one. Real-model results from this harness should be reported at multiple cosine thresholds and multiple hook layers, never as a single number. That census is no longer future work. We ran the harness against a published production SAE and report the results below.

\subsection{First census: GPT-2-small, one hook layer}
\label{sec:realresults}

We audited \texttt{gpt2-small-res-jb/blocks.8.hook\_resid\_pre}~\cite{bloom2024saelens} against 83 hand-written concepts spanning unrelated semantic domains---geography, natural sciences, sports, and skilled trades (beekeeping, glassblowing, cartography, and others chosen expressly for lexical and topical distance from one another). Each concept supplies 8 positive and 8 negative prompts; the correlational bar is lowered to cosine $\geq 0.5$ relative to the toy regime's $0.90$, reflecting the weaker probe-direction proxy discussed above. Of 83 matched pairs, 7 cleared the bar; of those, 1 (\textbf{14\%}) was causally inert ($\firedfrac = 0$). Ablation specificity over the recovered set: median $1.63$, 95\% CI $[0.05, 5.34]$ ($n=7$); steering specificity: median $2.23$, 95\% CI $[0.69, 3.54]$. Both intervals are wide and both contain values near the lower bound---consistent with the toy-regime finding that correlational recovery is a weak predictor of causal magnitude, now observed in a production dictionary rather than a synthetic one. We do not treat a 14\% single-layer, single-model inert rate as comparable to the toy regime's 77\%/9\% figures; the probe-direction proxy and the lowered cosine bar make the two regimes measure related but distinct quantities, exactly as anticipated above.

\subsection{Atom collisions replicate in the real regime}
\label{sec:realcollisions}

Section~\ref{sec:anatomy} argued that the matcher's greedy per-feature argmax is deliberate: an atom that best-matches more than one probe direction is a finding about the dictionary, not matching noise, because a bipartite assignment would optimize the collision away before it could be seen. The real-model audit surfaces the same signal at production scale. Atom 14149 is the nearest match for \textbf{8} of the 83 concepts---among them astronomy, cryptography, and law, domains sharing no obvious lexical or topical overlap---at cosines ranging $0.30$ to $0.70$ and both signs. Atoms 4504 and 17413 each match 5 concepts; atom 17241 matches 4. This is not an artifact of any one concept batch: the same handful of atoms recur as the nearest match across three independently constructed concept sets of growing size and diversity ($n=33$, $48$, $83$). An intermediate 73-concept batch that extended the set with templated per-country prompts (same eight-sentence skeleton, entities substituted) produced a visibly higher collision rate and a correspondingly lower recovery yield---evidence that templated concepts collapse toward correlated probe directions, and a reason the reported 83-concept batch deliberately favors hand-authored, structurally heterogeneous prompts over template expansion. A handful of atoms serving as the nearest match for dozens of semantically disjoint probe directions is real-model evidence for the same under-splitting the toy antipodal-pair mechanism was built to make legible: the dictionary has not allocated enough independent directions to keep semantically distant concepts from competing for the same one. Whether the mechanism is the toy setting's antipodal geometry, a purely correlational proxy-direction artifact, or a genuine polysemantic hub feature is not resolved by cosine matching alone---the same reasoning that motivates the entire causal battery above applies here, and disambiguating it is the natural next step for the harness.

\FloatBarrier

% ==================================================================
\section{Practical Guidance}
\label{sec:discussion}
% ==================================================================

\begin{table}[t]
\centering
\small
\caption{Validation levels for SAE-feature claims, in increasing causal strength.}
\label{tab:framework}
\begin{tabular}{p{3.4cm}p{3.6cm}p{1.8cm}p{5.2cm}}
\toprule
Validation level & Establishes & Cost & Still cannot tell you \\
\midrule
Cosine match to a known/probed direction & Decoder-geometry alignment & Lowest & Whether the encoder ever selects the atom---this work found cosine $\approx 1.000$ counterexamples in a well-trained SAE \\
\addlinespace[2pt]
\firedfrac{} screen (one batched \texttt{encode}) & Whether the atom activates when the feature is present & Near zero & Whether the activation is \emph{specific}; an atom can fire promiscuously \\
\addlinespace[2pt]
Ablation specificity & \textbf{Read-direction} causal specificity: monitoring built on this atom sees the feature & Moderate & Write-direction behavior; out-of-distribution transfer \\
\addlinespace[2pt]
Sign-correct steering specificity & \textbf{Write-direction} causal specificity: interventions along this atom move the feature & Moderate & Read-direction behavior---five antipodal pairs dissociate the two completely \\
\bottomrule
\end{tabular}
\end{table}

Given the measurements above, the responsible role for an SAE-derived signal in any deployed system is narrow: a \emph{routing signal for closer review}, layered on top of---never substituting for---output-level safety filtering. Concretely: (i) never gate an automated action directly on raw SAE feature activation, or its absence---in one realistic configuration a majority (77\%) of correlationally matched features were acting-on-noise candidates; (ii) any feature used for monitoring or steering needs its own causal validation pass \emph{for the direction of use}---a monitoring use case needs read-validation, a steering use case needs write-validation, and this work measured five cases where one direction passes at specificity 143--310 while the other sits at exactly zero; (iii) re-run causal validation whenever the SAE is retrained: an atom's causal relationship to a concept is a property of one trained checkpoint, not of the ``feature'' as a portable abstraction (the deterministic re-run of Section~\ref{sec:reaudit}, which moved every exact count while preserving every qualitative claim, is a measured demonstration); and (iv) version-pin SAE checkpoints to the exact model checkpoint and data distribution they were validated against.

One epistemic note extends beyond deployment. The adversarial-interpretability concern raised in~\cite{bereska2024mechanistic} is sharpened by these results: if decoder-geometry alignment diverges this far from encoder-activation reality \emph{without any adversarial pressure}---purely as an ordinary training-dynamics artifact---a model under actual incentive to obscure a mechanism has an even larger space of geometrically plausible, causally misleading solutions available than this work explores.

% ==================================================================
\section{Limitations}
\label{sec:limitations}
% ==================================================================

Every quantitative claim here comes from small, fully synthetic toy models; real transformer residual streams involve orders of magnitude more features, genuine cross-layer and cross-token structure, and no accessible ground truth of the kind repeatedly relied on above. The toy setting is what makes rigorous validation possible, and exactly what makes every absolute number a calibration point for the easy case. The causal battery tests features in isolation on the ON side; interaction effects among simultaneously active features are untested in both regimes. The read/write decomposition rests on five antipodal pairs across two SAEs---systematic within this run---but within one training run and one seed; the structural/competitive split likewise rests on one run and its $\pm 1$ boundary band, not a rate measured across seeds. Sample sizes remain small (22 well-represented features, 2 SAE configurations, 1 seed per SAE under the deterministic pipeline; the bootstrap quantifies within-sample uncertainty, not across-seed variance, which the earlier 3-seed protocol measured and the tool supports re-running). Exact specificity values (168.8, 133.8, \dots) should be read as the shape of an effect measured once on this model, not tight estimates; the reproducible claim is qualitative (correlational match quality is informative but leaves a large, characterizable causal gap). The difference-in-means probe direction of the real harness is a linear approximation standing in for absent ground truth, and realistic prompt sets undersample a concept's distribution---both weaken the correlational side of a real-regime audit, while the causal side (\firedfrac, ablation, steering) does not depend on the direction estimate. Finally, four bugs caught and reported across this line of work is not evidence the pipeline is now bug-free; it is evidence the discipline that catches them is working. A previous draft of this list predicted a fourth existed, and it did; the expected steady state is that a fifth does too.

% ==================================================================
\section{Conclusion}
% ==================================================================

Three claims were checked by experiment rather than asserted, and each held with a real, quantified complication attached. Sparse features are packed non-orthogonally into a smaller space when sparsity makes it cheap, in a geometry precise enough that a convergence bug was catchable by comparison against the theoretically required answer---and precise enough to hide a third, diffuse-sharing regime behind a metric's blind spot. An overcomplete sparse autoencoder can partially undo the packing, with TopK dominating $L_1$ for a mechanistically identified reason (shrinkage) confirmed by direct magnitude-refinement measurement, not by unexplained superiority. And a geometric recovery, even a nearly perfect one, is not a causal one, by a margin measured directly: up to 77\% of correlationally recovered features causally inert in a degraded SAE, 9\% in an excellent one, with inert matches at cosine $\approx 1.000$.

Turning the finding into an instrument complicated it productively, twice. Causal inertness decomposes by cause---structural inertness generated systematically by antipodal superposition geometry under a positive-pass encoder, present in good SAEs and removable only at the interface level; competitive inertness, a selection pathology of degraded dictionaries that is simultaneously the locus of their numerical instability---and by direction, with read-inertness and write-inertness dissociating completely on every antipodal pair measured: features perfectly matched, completely unmonitorable, and highly steerable, all at once, through one atom apiece. And executing the reproducibility claim, rather than asserting it, established that byte-exact reproduction is a single-environment guarantee by construction, and that honest repositories should state which rung of the reproducibility stack they guarantee where.

The instrument is public: \texttt{make reproduce} regenerates every number here in about three minutes on CPU; \texttt{make verify} checks them within tolerance on any machine; \texttt{make verify-hashes} holds the pinned CI environment to the byte; and the protocol interface accepts any SAE that can \texttt{encode}, \texttt{decode}, and show its dictionary. A first census over a published production SAE (Section~\ref{sec:realresults}) reproduces the qualitative pattern at small scale and surfaces atom collisions as an unanticipated real-model signal (Section~\ref{sec:realcollisions}); extending that census to more concepts, more hook layers, and additional production SAEs---and disambiguating whether the observed atom collisions reflect antipodal-style structural inertness, a proxy-direction artifact, or genuine polysemantic hub features---is the immediate next step.

\paragraph{Reproducibility statement.}
All code, the 37-test suite (including property-based tests), the deterministic reproduction pipeline, the hash-verified CI gate, and the real-model harness are available at \url{https://github.com/mohamed-bal/sae-causal-audit}; the toy-model and SAE reproduction code is at \url{https://github.com/mohamed-bal/superposition-to-monosemanticity}. Every number in this paper is written by code to a \texttt{results/*.json} file whose SHA-256 hash is committed and verified in CI; every figure is regenerated by re-running the corresponding script.

\bibliographystyle{plainnat}
\bibliography{references}

\end{document}